\documentclass[runningheads]{llncs}
\usepackage[T1]{fontenc}
\usepackage{graphicx}
\usepackage{color}
\usepackage[english]{babel}
\usepackage{tabularx} 
\usepackage{amsmath}  
\usepackage{graphicx} 
\usepackage[margin=1in,letterpaper]{geometry} 
\usepackage[final]{hyperref} 
\usepackage{booktabs} 

\usepackage{comment} 

\usepackage[dvipsnames]{xcolor} 

\usepackage{breakcites}
\usepackage{amsfonts}

\usepackage{tikz}
\usetikzlibrary{shapes.geometric, arrows,automata,positioning}
\tikzstyle{startstop} = [rectangle, rounded corners, minimum width=1cm, minimum height=1cm,text centered, draw=black, fill=red!30]
\tikzstyle{chance} = [circle, minimum width=1cm, minimum height=0.5cm,text centered, draw=black, fill=yellow!30]
\tikzstyle{io} = [trapezium, trapezium left angle=70, trapezium right angle=110, minimum width=0.5cm, minimum height=1cm, text centered, draw=black, fill=blue!30]
\tikzstyle{process} = [rectangle, minimum width=1cm, minimum height=1cm, text centered, draw=black, fill=orange!30]
\tikzstyle{decision} = [diamond, minimum width=1cm, minimum height=1cm, text centered, draw=black, fill=green!30]
\tikzstyle{arrow} = [thick,->,>=stealth]

\renewcommand{\vec}[1]{\underline{#1}}

\begin{document}
\title{Design and Implementation of an Heuristic-Enhanced Branch-and-Bound Solver for MILP 
\thanks{This work was funded by FRQ-IVADO Research Chair in Data Science for Combinatorial Game Theory, and NSERC grant 2019-04557. The authors express their gratitude to Sangnie Bhardwaj, Matteo Cacciola, Margarida Carvalho,  Gabriele Dragotto, Youssouf Emine, Maxime Gasse, Khalid Laaziri, Alessandra Lingua, Andrea Lodi, Christopher Morris, Krunal Patel, Mike Pieper, Jean-Yves Potvin,  Antoine Prouvost, Lea Ricard, Claudio Nazionale Sole and Mathieu Tanneau for the insightful conversations and comments.
A special thanks go to Prof. Carvalho and Prof. Lodi for their support.}}
%
%
\author{Warley Almeida Silva\inst{1} \and
Federico Bobbio\inst{1,2} \and
Flore Caye\inst{1} \and Defeng Liu\inst{2} \and Justine Pepin\inst{1} \and Carl Perreault-Lafleur\inst{1} \and William St-Arnaud\inst{1,3}}
\authorrunning{Margaridinhas team}
%
\institute{CIRRELT-DIRO, Université de Montréal, Montréal, Canada \email{\{warley.almeida.silva, federico.bobbio, fore.caye, justine.pepin.1, carl.perreault-lafleur, william.st-arnaud\}@umontreal.ca} \\
\and
CERC in Data Science, Polytechnique Montréal, Montréal, Canada
\email{defeng.liu@polymtl.ca}\\
 \and
MILA Institute, Montréal, Canada
}
\maketitle              
\begin{abstract}
We present a solver for Mixed Integer Programs (MIP) developed for the \emph{MIP competition 2022}. Given the 10 minutes bound on the computational time established by the rules of the competition, our method focuses on finding a feasible solution and improves it through a Branch-and-Bound algorithm. 
Another rule of the competition allows the use of up to 8 threads. Each thread is given a different primal heuristic, which has been tuned by hyper-parameters, to find a feasible solution. In every thread, once a feasible solution is found, we stop and we use a Branch-and-Bound method, embedded with local search heuristics, to ameliorate the incumbent solution. 
The three variants of the Diving heuristic that we implemented manage to find a feasible solution for 10 instances of the training data set. These heuristics are the best performing among the heuristics that we implemented. Our Branch-and-Bound algorithm is effective  on a small portion of the training data set, and it manages to find an incumbent feasible solution for an instance that we could not solve with the Diving heuristics. Overall, our combined methods, when implemented with extensive computational power, can solve 11 of the 19 problems of the training data set within the time limit.
Our submission to the MIP competition was awarded the "Outstanding Student Submission" honorable mention.

\keywords{MIP Competition 2022  \and Mixed Integer Linear Programming\and Primal Heuristics  \and Branch-and-Bound}
\end{abstract}
%
%
\section{Introduction}
\emph{Mixed-integer programming} (MIP) is a principal mathematical framework for modeling complex \emph{combinatorial optimization} (CO) problems, which has a wild range of applications, such as manufacturing, retail, scheduling and transportation, etc. Without loss of generality, MIP has the goal of finding optimal solutions with regard to some objective functions in the presence of constraints. In this paper, we focus on \emph{mixed-integer linear programs} (MILP) of the form
\begin{align*}
 (P)~~~\min~~~ &\boldsymbol{c}^T\boldsymbol{x}\\
\text{s.t.}~~~  &\boldsymbol{Ax} \le \boldsymbol{b},\\
& x_i \in \{0,1\}, ~~~\forall i\in \mathcal{B}, \\
& x_j \in \mathbb{Z}^+,  ~~~\forall j\in \mathcal{G},  \\
& x_k \ge 0, ~~~\forall k\in \mathcal{C}, 
\end{align*}
where the index set of decision variables $\mathcal{N}:=\{1,\ldots,n\}$ is partitioned into $\mathcal{B}, \mathcal{G}, \mathcal{C}$, which are the index sets of binary, general integer and continuous variables, respectively.

Over the last decades, there has been stunning progress in the field. Major developments of techniques to solve MILP problems include \emph{Branch-and-Bound} (B\&B) \cite{land2010automatic} and \emph{Cutting-Plane} \cite{Bixby2007} methods. These techniques have been successfully applied to address various problems from operations research as well as theoretical computer science. Needless to say, there is a rich scientific literature on MIP and CO \cite{korte2012combinatorial, grotschel2012geometric, wolsey2014integer}.

Due to the NP-hardness of many MIP problems, practical solvers incorporate a variety of complex algorithmic techniques, such as heuristics \cite{blum2003metaheuristics}, pre-processing \cite{araujo2020strong}, searching rules (often referred to as building blocks of the solution process) and result in complex and sophisticated software tools \cite{bonami2008algorithmic}. However, there are still many practical cases where existing techniques are not adequate and it is compelling to implement new MIP algorithms.

Primal heuristics \cite{fischetti2005feasibility, fischetti2003local, liu2021learning, gendreau2010handbook, fischetti2010heuristics} play a pivotal role to provide feasible solutions to global methods such as B\&B and \emph{Branch-and-Cut}, or even to start the search for local optima. Based on this context, the \emph{MIP Competition 2022} aims at creating novel general-purpose primal heuristics for MILP problems. In this paper, we propose \emph{a heuristic-enhanced B\&B based solver} for MILP, and summarize our solution for the MIP competition. Our submission to the MIP competition was awarded the "Outstanding Student Submission" honorable mention. 

In our investigation, we first focused on finding a feasible solution for the simple instances; then, we produced heuristics that would improve the incumbent solution and heuristics built \emph{ad hoc} for finding a feasible solution for the harder instances.

We implemented some well-known heuristics in the literature and then produced some original variants of these. As a result of our work, we have found that the \emph{Diving}\cite{berthold2006primal} heuristics perform very well compared to the other primal heuristics, finding a feasible solution for 10 out of 19 problems within the 10 minutes time limit. The \emph{Feasibility Pump} \cite{fischetti2005feasibility} could find a feasible solution for few instances; in all these instances the Diving heuristics could find better incumbents. We also implemented \emph{Repairing Local Branching} \cite{RLB2006.08.004}, warmed up by the Feasibility Pump or the Diving; our implementation of the Repairing Local Branching warmed up by the Feasibility Pump could find a solution to a problem that the Divings could not solve. 
At the beginning of our work, we also implemented methods based on  \emph{Simulated Annealing} and \emph{Genetic algorithms} \cite{gendreau2010handbook}, but they performed poorly, hence we did not include them in this report. We also worked on the \emph{Lagrangian Relaxation} \cite{fisher1981lagrangian}. We created new variants (up to our knowledge) of some heuristics: 

\begin{itemize}
    \item we implemented learning on the Diving heuristics (in a different fashion from \cite{fortin2021diving}) and other primal heuristics,
    
    \item we explored the effects of substituting Diving to the Feasibility Pump in the Repairing Local Branching,
\end{itemize}

In none of these cases we could achieve a performance that would outperform the simple Divings for finding a feasible initial solution. All the research we have conducted has opened up many promising paths that we plan to investigate in the near future.

By using our different heuristics, we manage to find feasible solutions for 11 problems out of 19 with extensive computational resources and within 10 minutes. We build a pipeline method, called \textit{Poutine}, that implements the best performing heuristics in parallel and then refines each feasible solution through Branch-and-Bound; at the end of the time-limit, the method outputs the best incumbent. When the computational power and the time-limit are set according to the constraints of the competition, our general pipeline method can only find feasible solution to few problems. Therefore, we restrict the parallel computations and allow to run only the Diving heuristics; nonetheless, we still could not distribute the resources to solve the same amount of public instances efficiently in parallel. We plan to ameliorate our methods and explore more thoroughly the many research directions we examined. 

The remaining part of the paper is organized as follows. In the next section, we introduce the MIP competititon 2022 and analysis of the datasets. In section \ref{sec: method}, we present the overview of our method. In section~\ref{sec: primal heuristics}, we present the primal heuristics that we developed. Then, in section~\ref{sec: local heuristics}, we present the local heuristics for enhancing B\&B. Finally, in section~\ref{sec: results}, we analyse the outcome of our experiments and we draw the conclusions.

\section{The MIP competition 2022}
\label{sec:compettion}

The competition is divided in two stages. In the first stage, the proposed methods are evaluated on their novelty and on a public data set of 19 MIP instances drawn from the benchmark set of MIPLIB 2017 \cite{gleixner2021miplib}. In the second step, the winner of the competition is established by testing the selected method on a hidden test data set.

The training set of the competition is made of 19 instances, two of which are classified as \emph{hard} by MIPLIB. We ran Gurobi with a time-limit of 600 seconds on 8 threads on the server of Compute Canada with 16GB of RAM and a 2.4 GHz Intel Gold 6148 Skylake. We used this machine also for the other experiments that we present in this document. The programming language Julia (version 1.7) was chosen for the implementation of our method.
In Table~\ref{Table: preliminary-analysis}, we report the results of the preliminary investigation about Gurobi's performance on the public training data set. These results will be a benchmark for our heuristics. First of all, it is relevant to notice that the presolve is very quick in all instances, and for some of them it reorganizes dramatically the number of constraints and variables. The instances solved to optimality within the time-limit consist of 14 of the 19 in the public data set. It is worth mentioning that both instances highschool1-aigio  and supportcase10 are classified as hard by the MIPLIB 2017: Gurobi struggled on the former, and solved to optimality the latter. Overall, Gurobi could not find an incumbent solution for highschool1-aigio, cryptanalysiskb128n5obj16, and neos-1354092 within the 10 minutes time-limit. As pointed out in \cite{applegate2021practical}, solving the LP of highschool1-aigio required 70 seconds, corroborating our findings. 

\begin{table}[ht]
\centering
\begin{tabular}{lrrrrrrrrr}
 \hline
  Instance  &  Pr-T   & opt-T & \#nodes   & Val-inc & Exact &   const & bin & int & cont  \\  
  \hline
  
 academictimetablesmall &	0.26 & \textcolor{BrickRed}{$>$600.00} & 7952 & 1.00 &	0.00 &		23294 &	28926	&	0 &	 0 \\ 
 
comp07-2idx	&	0.13 & 54.93 & 729 & 6.00 &	6.00 &		21235 &	17155	&	109	&	0 \\ 

cryptanalysiskb128n5obj16 & 0.67 & \textcolor{BrickRed}{$>$600.00} & 3 & \textcolor{BrickRed}{\textbf{NULL}} & 0.00 &  98021 & 47830 & 1120 & 0 \\  

eil33-2 	&	0.04 & 1.32 & 4733 & 934.01 &	934.01		&	32	&	4516	&	0	&	0 \\  

highschool1-aigio & 0.19 & \textcolor{BrickRed}{$>$600.00} & 1 & \textcolor{BrickRed}{\textbf{NULL}} & 0.00 	&	92568	&	319686	&	718	&	0 \\  
 
mcsched 	&	0.01 & 8.33 & 11390 & 211913.00  &	211913.00 &			2107	&	1745	&	0	&	2 \\  

neos-1354092	&	0.12 & \textcolor{BrickRed}{$>$600.00} & 7875 & \textcolor{BrickRed}{\textbf{NULL}} &	46.00 &			3135	&	13282	&	420	&	0 \\  

neos-3024952-loue 	&	0.01 & 600.01 & 1947175 & 26756.00 &	26756.00 &			3705	&	0	&	3255	&	0 \\  

neos-3555904-turama	&	1.97 & 108.56 & 1 & -34.70 &	-34.70 &			146493	&	37461	&	0 &	0 \\  

neos-4532248-waihi	&	1.55 & 119.47 & 588 & 61.60 &	61.60	&	167322	&	86841	&	0	&	1 \\  

neos-4722843-widden 	&	0.96 & 38.22 & 1 & 25009.66 &	25009.66 &			113555	&	73349	&	20	&	4354 \\  

ns1760995	&20.99 & 498.88 & 1 & -549.21 &	-549.21	&	615388	&	17822	&	0	&	134 \\  

ns1952667	&	0.33 & 42.19 & 12431 & 0.00 &	0.00 &		41	&	0	&	13264	&	0 \\  

peg-solitaire-a3 	&	0.03 & 178.13 & 1833 & 1.00 &	1.00 	&	4587	&	4552	&	0	&	0 \\  

qap10 	&	0.03 & 20.05 & 1 & 340.00  &	340.00 &			1820	&	4150	&	0	&	0 \\  

rail01 	&	0.50 & 155.7 & 1 & -70.57 &	-70.57		&	46843	&	117527	&	0	&	0 \\  

rococoC10-001000 & 0.01 & 52.02 & 24307 & 11460.00 &	11460.00 & 1293 & 2993 & 124	& 0 \\  

seymour &	0.03 & \textcolor{BrickRed}{$>$600.00} & 32461 & 424.00 &	423.00 	&	4944	&	1372	&	0	&	0 \\  

supportcase10 & 0.69 & 549.26 & 450 & 7.00 &	7.00 & 165684 & 14770 & 0	& 0 \\\hline

\end{tabular}
\caption{The table reports the preliminary study on the training data set conducted with Gurobi (time-limit 600 seconds). Key: Pr-T is the presolve time of Gurobi; opt-T is the time required to find the  optimal solution; \#nodes is the number of explored nodes; Val-inc is the value of the best incumbent solution ("NULL" means that no incumbent was found); Exact is the value of the optimal solution; const is the number of constraints; bin is the number of binary variables; int is the number of general integer variables; con is the number of continuous variables.}
\label{Table: preliminary-analysis} 
\end{table}

\section{Method}\label{sec: method}

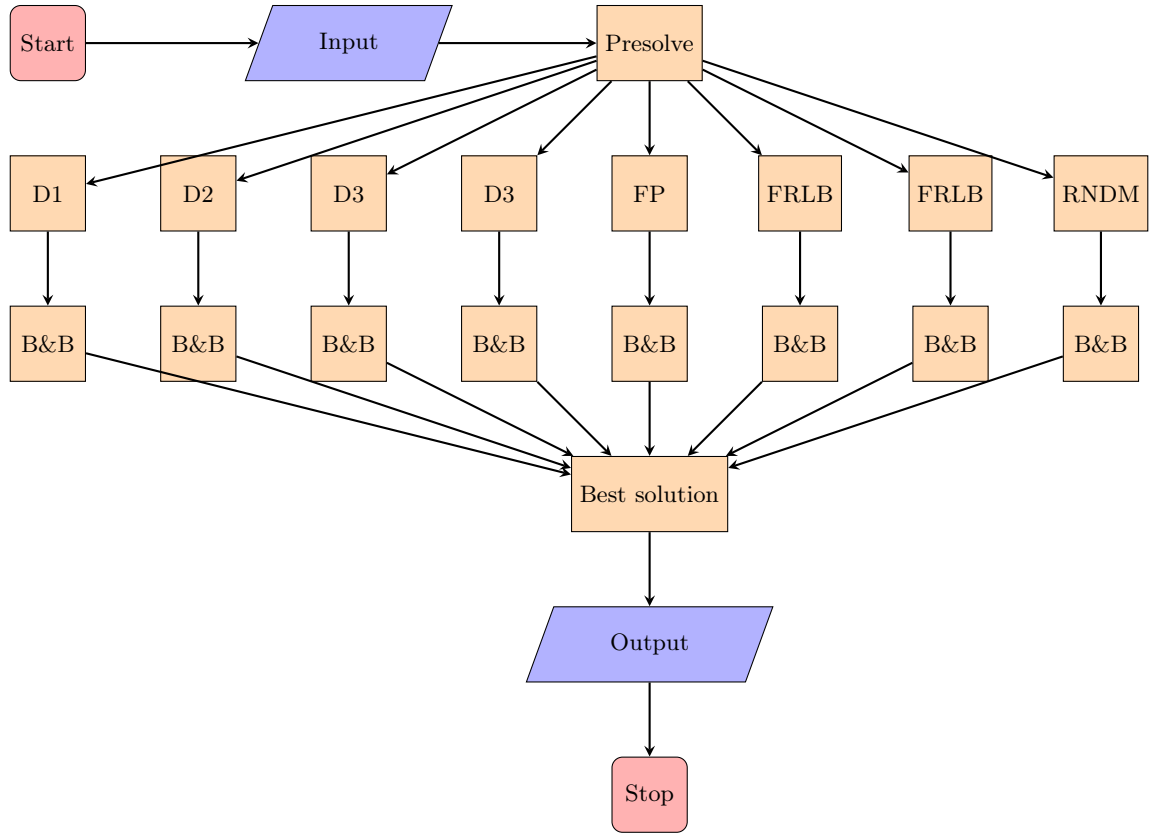
\begin{figure}[ht]
\centering  
  \begin{tikzpicture}[node distance=2.0cm,
on grid,
    auto]
\node (start) [startstop] {Start};
\node (in1) [io, right of=start, xshift=2cm] {Input};
\node (pro1) [process, right of=in1, xshift=2cm] {Presolve};

\node (Heur5) [process, below of=pro1] {FP}; 
\node (Heur4) [process, left of=Heur5] {D3};
\node (Heur3) [process, left of=Heur4] {D3};
\node (Heur2) [process, left of=Heur3] {D2};
\node (Heur1) [process, left of=Heur2] {D1};

\node (Heur6) [process, right of=Heur5] {FRLB}; 
\node (Heur7) [process, right of=Heur6] {FRLB}; 
\node (Heur8) [process, right of=Heur7] {RNDM};

\node (Improve1) [process, below of=Heur1] {B\&B};
\node (Improve2) [process, below of=Heur2] {B\&B};
\node (Improve3) [process, below of=Heur3] {B\&B};
\node (Improve4) [process, below of=Heur4] {B\&B};
\node (Improve5) [process, below of=Heur5] {B\&B};
\node (Improve6) [process, below of=Heur6] {B\&B};
\node (Improve7) [process, below of=Heur7] {B\&B};
\node (Improve8) [process, below of=Heur8] {B\&B};

\node (Best) [process, below of=Improve5] {Best solution};

\node (out1) [io, below of=Best] {Output};
\node (stop) [startstop, below of=out1] {Stop};

\draw [arrow] (start) -- (in1);
\draw [arrow] (in1) -- (pro1);

\draw [arrow] (pro1) -- (Heur1);
\draw [arrow] (pro1) -- (Heur2);
\draw [arrow] (pro1) -- (Heur3);
\draw [arrow] (pro1) -- (Heur4);
\draw [arrow] (pro1) -- (Heur5);
\draw [arrow] (pro1) -- (Heur6);
\draw [arrow] (pro1) -- (Heur7);
\draw [arrow] (pro1) -- (Heur8);

\draw [arrow] (Heur1) -- (Improve1);
\draw [arrow] (Heur2) -- (Improve2);
\draw [arrow] (Heur3) -- (Improve3);
\draw [arrow] (Heur4) -- (Improve4);
\draw [arrow] (Heur5) -- (Improve5);
\draw [arrow] (Heur6) -- (Improve6);
\draw [arrow] (Heur7) -- (Improve7);
\draw [arrow] (Heur8) -- (Improve8);

\draw [arrow] (Improve1) -- (Best);
\draw [arrow] (Improve2) -- (Best);
\draw [arrow] (Improve3) -- (Best);
\draw [arrow] (Improve4) -- (Best);
\draw [arrow] (Improve5) -- (Best);
\draw [arrow] (Improve6) -- (Best);
\draw [arrow] (Improve7) -- (Best);
\draw [arrow] (Improve8) -- (Best);

\draw [arrow] (Best) -- (out1);

\draw [arrow] (out1) -- (stop);
\end{tikzpicture}
\caption{Description tree of the method: We input the instance model. We presolve the model. We run 8 primal heuristics in parallel to find a feasible solution. As soon as a primal heuristic finds a feasible solution, we improve it with a Branch-and-Bound-based method until the time-limit is reached. The incumbent solution is written to the .sol output file as it is found. The eight primal heuristics are variants of: Diving (D1, D2, D3), Feasibility Pump (FP) and Repairing MIP for Local Branching with FP (FRLB).}
\label{Figure: Description_tree}
\end{figure}

Our method is composed of the following steps.
\begin{enumerate}
    \item Presolve the instance.
    \item  Create 8 threads to run different heuristics, by type or by parameters. In parallel, in each thread:
    \begin{enumerate}
        \item Run the primal heuristic.
        \item If a feasible solution was found and there is still some time, improve through B\&B using the incumbent solution as a warm-up. 
    \end{enumerate}
\end{enumerate}

In Figure~\ref{Figure: Description_tree} we represent the description tree of our method. 
The input consists in uploading the MIP instance, that we immediately simplify through the presolve. Once we have obtained our presolved model, we save it and we provide it to 8 primal heuristics. Each heuristic is allocated a thread with its own memory, and they all work in parallel on finding a feasible solution. In every thread, if a heuristic finds a feasible solution within the 10 minutes time-limit, we pursue a better incumbent through a Branch-and-Bound based method. The incumbent solution is shared through a piece of memory protected with locks, and transferred to the final output file as it is updated.

One of the main results of our contribution is a breadth-first search Branch-and-Bound algorithm, that we use as a MIP solver for both local and global search methods.  

In the next section we introduce the heuristics that we have produced and describe in detail how we select them. 

We have also explored primal methods based on  Simulated Annealing and Genetic algorithms \cite{gendreau2010handbook}, these methods did not produce satisfactory results.

\section{Primal heuristics}\label{sec: primal heuristics}

Taking inspiration from the literature~\cite{achterberg2007constraint,conforti2014integer,berthold2015heuristic}, we produced several versions of the following primal heuristics: Diving \cite{berthold2006primal}, Feasibility Pump \cite{fischetti2005feasibility,berthold2019ten}, 
Repairing Local Branching \cite{RLB2006.08.004}. Next, we present our implementation of these heuristics; the experimental results we obtained are presented in Section~\ref{sec: results}. 

\paragraph{Diving}
This heuristic starts by finding a solution to the continuous relaxation of the problem. If the solution is integer feasible, we stop. Otherwise, as the name suggests, we proceed by bounding one integer constrained variable (that has non-integer value) and solving the obtained continuously relaxed sub-problem until all the non-integer valued variables are bounded. Once we reach the leaf of the branch, if the solution is not integer feasible, we go back in the branch and remove all the bounds until we reach a node that has still some open branches. 

Once we obtain a non-feasible solution, there are several ways of choosing the variable to bound. We produce three variations of the Diving heuristic on the basis of different selection methods. 

\begin{itemize}
    \item Dive-1: We select the variable with the smallest gap from an integer (indifferently, upper or lower rounding).
    \item Dive-2: We select the variable with the greates gap from an integer (indifferently, upper or lower rounding).
    \item Dive-3: We select the variable randomly.
\end{itemize}

\paragraph{Feasibility Pump} 
The Feasibility Pump (FP) starts by finding a feasible solution to the continuously relaxed MIP problem. If the solution found is not integer feasible, then we round it. If the rounded solution is not feasible, we proceed to the following loop: We compute the solution to the continuously relaxed problem with the same constraints as the original problem and with the Hamming distance from the last rounded solution as the objective. If the solution found is not integer feasible, then we round it and we find a new solution. To break loopholes, the algorithm introduces some perturbations.  

Since the introduction of the Feasibility Pump (FP), many researchers have explored ways to improve it \cite{berthold2019ten}. We implement a variation\footnote{In the denominator of the right addend we add +1 to avoid division by 0 when $c=\vec{0}$. Indeed, the instance cryptanalysiskb128n5obj16 has exactly such a cost vector.} of the FP that introduces a convexification term in the objective function \cite{achterberg2007improving}:

\begin{equation}\label{Formula: Feasibility Pump}
 \Delta_\alpha(x,\Tilde{x}) = (1-\alpha)\Delta(x,\Tilde{x}) +\alpha \left(\frac{\sqrt{|\mathcal{I}|}}{\lVert c \rVert +I(\lVert c \rVert=0)}\right)  c^\top x,
\end{equation}

where $\Tilde{x}$ is the rounded vector of the last feasible relaxed solution, $\alpha\in [0,1]$, $\Delta(x, \Tilde{x})$ is the Hamming distance between $x$ and $\Tilde{x}$, $\mathcal{I}$ is the set of integer variables, and $c$ is the cost vector of the original model. The convex parameter $\alpha$ will be useful for hyper-parameter tuning purposes on the training set. 
Our FP heuristics are based on the best performing values of $\alpha$.

\paragraph{Repairing MIP for Local Branching}
Local Branching \cite{fischetti2003local} (LB) is a local search method that yields a branching criterion within an enumerative scheme for the problem under investigation.
Given an integer feasible solution, the method first defines a series of LB neighborhoods, then explores them by calling a black-box MIP solver. More details about LB can be found in Section \ref{sec: local heuristics}.
Repairing MIP for Local Branching (RLB) is a modification of LB that starts with a integer vector $\Hat{x}$ which is near the feasibility region; we find such initial vector $\Hat{x}$ by running some primal heuristic like FP for some iterations, typically 10 to 100. Once we have selected $\Hat{x}$, we add an artificial continuous variable to every constraint violated by $\Hat{x}$. We also add constraints on the artificial continuous variables that forces an artificial binary variable to be bigger than zero when the artificial continuous variable is not set to zero. Thus, each artificial binary variable serves as a flag to indicate if we are using a slack variable or not. Then, we change the objective to the sum of the artificial binary variables, and we try to minimize that objective. When it reaches 0, it means the RLB has found a feasible solution to the original problem. When we use as a primal heuristic FP, we obtain Feasibility Repairing Local Branching (FRLB). When we use as a primal heuristic Diving, we obtain Diving Repairing Local Branching (DRLB).

\section{B\&B-based heuristics}\label{sec: local heuristics}

As we mentioned, we implemented our own version of B\&B, that we use both as a MIP solver and as a global search method. To improve the computation of B\&B, we implemented some well-known neighborhood search heuristics: \emph{Local Branching} (LB) \cite{fischetti2003local}, Large Neighborhood Search (LNS) \cite{shaw1998using,pisinger2010large} and Relaxation Induced Neighborhood Search (RINS) \cite{danna2005exploring}. 

\paragraph{LB}
The LB heuristic and its variants have been the first primal heuristics exploring variable neighborhoods through off-the-shelf MIP solvers. It was proposed to improve the incumbent solution at a very early stage of the computation. Indeed, computing high-quality early solutions can improve the primal bound and reduce the size of the Branch-and-Bound tree by pruning more nodes. Given a feasible solution, the method first defines a solution neighborhood through the so-called \emph{Local Branching constraint}, namely, a linear inequality limiting the Hamming distance from the incumbent. 

Specifically, given the incumbent solution $\bar{x}$, let $\bar x_j$ denote the value of the feasible solution for binary variable $x_j, j\in \mathcal{B}$, where $\mathcal{B}$ denotes the index set of the binary variables; we also denote by $\mathcal{S} = \{ j\in \mathcal{B} : \bar x_j=1 \}$ the binary support of $\bar{x}$. For a given positive integer parameter $k$, we define the Local Branching neighborhood $N(\hat{x}, k)$ as the set of feasible solutions of the new MIP.
The new MIP that we define has the following additional constraints, known as \emph{Local Branching constraint}:

\begin{equation}
\label{eq_phi}
\Delta(x, \bar{x}) = \sum_{j\in \mathcal{B}\setminus \mathcal{S} }x_j + \sum_{j\in \mathcal{S}}(1-x_j) \le k.
\end{equation}

The algorithm iteratively explores a sequence of LB neighborhoods by calling the underlying MIP solver with the given computing resources (time limit or B\&B node limit).

\paragraph{LNS}
LNS is a large neighborhood search heuristic. Given as an input a feasible solution $\Bar{x}$, it searches the best feasible solution in the neighbourhood of $\Bar{x}$ (the size of the neighborhood is a parameter). Once the best feasible solution $\Tilde{x}$ in the neighborhood is found, the procedure updates $\Bar{x}= \Tilde{x}$. The method keeps searching for the best feasible solution in the new neighborhood until the stopping criterion is reached.

\paragraph{RINS}
RINS is another large neighborhood search heuristic that leverages the gap between the incumbent $\Bar{x}$ and the feasible solution of the continuous relaxation $x_R$ of the current node in the B\&B tree. It consists of three steps: 1) Fix the variables that have the same value in $\Bar{x}$ and $x_R$; 2) Set an objective cutoff based on the objective value of $\Bar{x}$; 3) Solve the obtained sub-MIP.

\section{Results}\label{sec: results}

In this section we present the experimental results on the primal heuristics, then on the Branch-and-Bound, and finally on the pipeline method. 

\paragraph{Experiments to select the primal heuristics}    

\begin{table}
\hspace*{-8pt}\makebox[\linewidth][c]{
\begin{tabular}{lrrrrrr}
\toprule
                  Instance &         Exact &    Gurobi &  Dive-1 &  Dive-2 &  Dive-3 & FRLB\_best  \\
\midrule
    academictimetablesmall &          0.00 &      1.00 &         - &         - &         - &         -  \\
               comp07-2idx &          6.00 &      6.00 &         - &    114.00 &    136.00 &         18568.00 \\
 cryptanalysiskb128n5obj16 &          0.00 &         - &         - &         - &         - &         - \\
                   eil33-2 &        934.01 &    934.01 &   1705.84 &   1218.09 &   2247.93 &         1000.24 \\
         highschool1-aigio &          0.00 &         - &         - &         - &         - &         - \\
                   mcsched &     211913.00 & 211913.00 & 267672.00 & 237847.00 & 221856.00 & 370383.00  \\
              neos-1354092 &         46.00 &         - &         - &         - &         - &         -  \\
         neos-3024952-loue &      26756.00 &  26756.00 &  34106.00 &  36565.00 &  33393.00 &         - \\
       neos-3555904-turama &        -34.70 &    -34.70 &    -33.20 &         - &         - &         -  \\
        neos-4532248-waihi &         61.60 &     61.60 &         - &         - &         - &         -  \\
       neos-4722843-widden & 25009.66 &  25009.66 &         - &  31696.71 &  31514.47 &         -  \\
                 ns1760995 &       -549.21 &   -549.21 &         - &   -119.24 &         - &         - \\
                 ns1952667 &          0.00 &      0.00 &         - &         - &         - &         -  \\
          peg-solitaire-a3 &          1.00 &      1.00 &         - &         - &         - &         - \\
                     qap10 &        340.00 &    340.00 &    558.00 &    402.00 &    464.00 &    340.00  \\
                    rail01 &        -70.57 &    -70.57 &         - &         - &         - &         - \\
          rococoC10-001000 &      11460.00 &  11460.00 &         - &         - &         - &        262689.00 \\
                   seymour &        423.00 &    424.00 &    440.00 &    439.00 &    435.00 &    532.00  \\
             supportcase10 &          7.00 &      7.00 &         - &     16.00 &     17.00 &         -  \\
\bottomrule
\end{tabular}
}\caption{We report the results of the best performing primal heuristics with a computational time-limit of 10 minutes.}
\label{Table: primal heuristics results}
\end{table}

We ran experiments to evaluate the primal heuristics that we produced on the public instances of the competition. 

The three Diving heuristics have a parameter that establishes the upper bound over the number of Divings we wish to perform. Given the characteristics of the competition, we set this parameter equal to a very high number, because we want the heuristic to find a feasible solution until the very last second. 

The Feasibility Pump has three parameters: $nT$ is the number of iterations to be done that we set as a very high number; $FP_T$ is the interval of the number of iterations at the end of which we perturb the variables to avoid being stuck with an infeasible integer solution; $\alpha$ is the convexification parameter of equation~\eqref{Formula: Feasibility Pump}.

The parameters of Local Branching are $k$, the size of the neighborhood, and $t$, the time-limit. Feasibility Repairing Local Branching uses the parameters of Feasibility Pump; similarly, Diving Repairing Local Branching uses the parameters of Diving.

The experiments are run in Julia 1.7 \cite{2106.08777} on the Cedar cluster of Compute Canada. In Table~\ref{Table: primal heuristics results} we present the results of the best performing primal heuristics. 

In general, the three Diving heuristics managed to find a solution for 10 instances. The heuristic Dive-2 finds the highest amount of feasible solutions. Note that the three Divings find complementary solutions, therefore we will need to implement the three of them in the final method.

The Feasibility Pump (results not reported in the table), was able to solve qap10 to optimality and seymour with a value of 541. The best performing parameter $\alpha$ was 0.4. Since FP was able to solve at least one instance to optimality, we will include it in our general method.

Given the overall unsatisfactory results of the Feasibility Pump, we implemented FRLB, i.e., we repair the likely unfeasible output of the FB after some iterations. In Table~\ref{Table: primal heuristics results} we report the best upper bounds found in 10 minutes by a certain variant of FRLB. Recall that in FRLB the parameters are the initial $\alpha$, the number of iterations (that we set to 10) and the number of perturbations (that we set to 20). Surprisingly, FRLB manages to find an incumbent feasible solution for an instance, rococoC10-001000, that none of the Divings could solve. DRLB achieved poorer results than FRLB.  

\paragraph{Experiments on Branch-and-Bound}
In Table~\ref{Table: Branch-and-bound} we present the results obtained from running the B\&B algorithm on the 19 instances with a time-limit of 10 minutes. The experiments are run in Julia 1.7 \cite{2106.08777} on the Cedar cluster of Compute Canada. As we can see, it finds a feasible solution to 5 instances. Note that B\&B finds a feasible solution to rococoC10-001000, for which the three Diving heuristics did not manage to find any. The B\&B was implemented with LB only at the root node and LNS and RINS alternatively every 50 nodes of the search tree. These local search heuristics proved to speed up the computation considerably. 

\begin{table}
\hspace*{-8pt}\makebox[\linewidth][c]{
\begin{tabular}{lrrr}
\toprule
                  Instance &         Exact &    Gurobi &   B\&B  \\
\midrule
    academictimetablesmall &          0.00 &      1.00 &           -  \\
               comp07-2idx &          6.00 &      6.00 &                 5457.00  \\
 cryptanalysiskb128n5obj16 &          0.00 &         - &            - \\
                   eil33-2 &        934.01 &    934.01 &           993.61 \\
         highschool1-aigio &          0.00 &         - &          - \\
                   mcsched &     211913.00 & 211913.00 & 220810.00  \\
              neos-1354092 &         46.00 &            - &         -  \\
         neos-3024952-loue &      26756.00 &  26756.00 &          - \\
       neos-3555904-turama &        -34.70 &    -34.70 &         -  \\
        neos-4532248-waihi &         61.60 &     61.60 &                  -  \\
       neos-4722843-widden & 25009.66 &  25009.66 &                 -  \\
                 ns1760995 &       -549.21 &   -549.21 &                  - \\
                 ns1952667 &          0.00 &      0.00 &                 -  \\
          peg-solitaire-a3 &          1.00 &      1.00 &               - \\
                     qap10 &        340.00 &    340.00 &       340.00  \\
                    rail01 &        -70.57 &    -70.57 &               - \\
          rococoC10-001000 &      11460.00 &  11460.00 &              148173.00  \\
                   seymour &        423.00 &    424.00 &       -  \\
             supportcase10 &          7.00 &      7.00 &                 -\\  
\bottomrule
\end{tabular}
}
\caption{We report the results of the Branch-and-Bound on the instances with a time-limit of 10 minutes.}
\label{Table: Branch-and-bound}
\end{table}

\paragraph{Experimental results on the pipeline}

We implemented our method, \emph{Poutine}, by doing first the presolve, and then solving the presolved model in each of the 8 threads. We choose the following methods in each thread with a time-limit of 10\textquotesingle:

\begin{enumerate}
    \item Dive-1, and then B\&B,
    \item Dive-2, and then B\&B,
    \item Dive-3, with seed 100, and then RLB, and then B\&B,
    \item Dive-3, with seed 200, and then B\&B,
    \item FP with $(nT = 10, FP_T = 20, \alpha = .4)  $ and then B\&B,
    \item FP with $(nT = 10, FP_T = 20, \alpha = .9) $ and then RLB, and then B\&B,
    \item FP with $(nT = 10, FP_T = 20, \alpha = .4) $ and then RLB, and then B\&B,
    \item Random instantiation of the variables to find a feasible solution. 
    
\end{enumerate}

Note that in the last thread, we run random instantiations of the model variables for 10 minutes in order to find a feasible solution. 

The experiments are run in Julia 1.7 \cite{2106.08777} on the Cedar cluster of Compute Canada. Table~\ref{Table: Poutine method} presents our results.
Since the computational limitation set by the competition cut the performance of our method by half, during the competition we restrict our threads to the first 4 in order to exploit the scarce computational resources. We still obtained similar computational results.

\begin{table}
\hspace*{-8pt}\makebox[\linewidth][c]{
\begin{tabular}{lrrrr}
\toprule
                  Instance &         Exact &    Gurobi &   \quad \emph{Poutine}-ub & \quad \emph{Poutine}-lb  \\
\midrule
    academictimetablesmall &          0.00 &      1.00 &           -  & 0 \\
               comp07-2idx &          6.00 &      6.00 &      114.00    & 0.00  \\
 cryptanalysiskb128n5obj16 &          0.00 &         - &            - & 0.00 \\
                   eil33-2 &        934.01 &    934.01 &          993.61 & 865.09 \\
         highschool1-aigio &          0.00 &         - &          - & 2.32*$10^{-12}$ \\
                   mcsched &     211913.00 & 211913.00 & 216891.00 & 200767.56  \\
              neos-1354092 &         46.00 &            - &         - & 18.98  \\
         neos-3024952-loue &      26756.00 &  26756.00 &          - & - \\
       neos-3555904-turama &        -34.70 &    -34.70 &         - & -  \\
        neos-4532248-waihi &         61.60 &     61.60 &                  - & -  \\
       neos-4722843-widden & 25009.66 &  25009.66 &                 - & -  \\
                 ns1760995 &       -549.21 &   -549.21 &        - & - \\
                 ns1952667 &          0.00 &      0.00 &           -  & 0.00      \\
          peg-solitaire-a3 &          1.00 &      1.00 &           -    & 1.00 \\
                     qap10 &        340.00 &    340.00 &       340.00 &       340.00 \\
                    rail01 &        -70.57 &    -70.57 &    -    & -90.76 \\
          rococoC10-001000 &      11460.00 &  11460.00 &    148173.00     &  7516.07 \\
                   seymour &        423.00 &    424.00 &       431.00 & 408.66 \\
             supportcase10 &          7.00 &      7.00 &                 - & 3.38 \\  
\bottomrule
\end{tabular}
}
\caption{We report the results of the \emph{Poutine} method on the public instances with a time-limit of 10 minutes.}
\label{Table: Poutine method}
\end{table}

\section{Conclusions}\label{sec: conclusions}
In this paper we have presented the methods that we explored for our submission to the MIP 2022 competition. Our submission to the MIP competition was awarded the "Outstanding Student Submission" honorable mention. 

The algorithms that we implemented range from standard ones to original variations of them. Concerning primal heuristic methods, we implemented methods based on Simulated Annealing, Genetic algorithms, Diving heuristics, Feasibility Pump, Repairing Local Branching, Branch-and-Bound. 

Moreover, we also produced neighborhood search heuristics to improve the quality of the incumbent feasible solution: Local Branching, Large Neighborhood Search and Relaxation Induced Neighborhood Search. 

By running the primal heuristics in parallel, we can find a feasible solution for 11 of the 19 problems of the training set of the competition within 10 minutes. This is a good result since Gurobi could find an initial feasible solution for 16 of the 19 problems. When we run our method with the computational constraints of the competition, i.e. the 16GB of RAM, then our method loses its computational efficiency and is able to solve fewer problems from the original training data set. 

Concerning the original contribution to the literature, we have explored:

\begin{itemize}
    \item Learning on Diving.
    \item Repairing Local Branching based on Diving.
\end{itemize}

For this project, we have written more than 3000 lines of code in Julia. 
Through our study, we have observed a remarkable efficiency of the Diving heuristic and we plan to explore the insights we gained in a future work. As shown by \cite{fortin2021diving, liu2021learning}, this is indeed a promising line of research.

%
%
%
 \bibliographystyle{splncs04}
 \bibliography{bibfile}

\end{document}